\documentclass[journal]{IEEEtran}
\usepackage{cite}
\usepackage{amsmath,amssymb,amsfonts}
\usepackage{algorithmic}
\usepackage{graphicx}
\usepackage{algorithm,algorithmic}
\usepackage{hyperref}
\hypersetup{hidelinks}
\usepackage{textcomp}
\usepackage{multirow}
\usepackage[caption=false]{subfig}
\usepackage{orcidlink}
\usepackage{balance}

\def\BibTeX{{\rm B\kern-.05em{\sc i\kern-.025em b}\kern-.08em
    T\kern-.1667em\lower.7ex\hbox{E}\kern-.125emX}}

\usepackage{tikz}
\usetikzlibrary{positioning, calc, arrows.meta}

\begin{document}

\title{A Dual Cross-Attention Framework for Colposcopic CIN Grading and Swede Score Prediction Using a New Multi-Center Dataset}

\author{
    Dania Khan$^{*}$, Nuzhat Aisha Shaikh$^{*}$, Asfina Hassan Juicy, Raiyun Kabir, \\S M Shahida, Taufiq Hasan \orcidlink{0000-0002-6142-3344},\IEEEmembership{Senior Member, IEEE} \thanks{D. Khan, N. A. Shaikh, A. H. Juicy, R. Kabir, and T. Hasan are with the mHealth lab, Department of Biomedical Engineering, Bangladesh University of Engineering and Technology (BUET), Dhaka, Bangladesh (e-mail:2018036@bme.buet.ac.bd; 2018022@bme.buet.ac.bd). T. Hasan is also with the Center for Bioengineering Innovation and Design (CBID), Johns Hopkins University, Baltimore, Maryland, USA. (e-mail:taufiq@bme.buet.ac.bd) S. M. Shahida is with Dhaka Medical College, Dhaka, Bangladesh.}
}

\maketitle
\begin{abstract}
Cervical cancer is a major global health challenge, with disease burden falling disproportionately on low- and middle-income countries (LMICs) due to a shortage of trained specialists and the subjective nature of colposcopy-based screening. To address this challenge, we propose a novel deep learning framework for the automated grading of Cervical Intraepithelial Neoplasia (CIN) and the prediction of clinical Swede scores. We also introduce the BUET Multi-Center Colposcopy Dataset, a novel, multi-center cohort designed and annotated for Swede score prediction and CIN grading. 
Our proposed dual-stream cross-attention architecture mimics the visual reasoning of an expert colposcopist by explicitly fusing paired multimodal cervigrams to evaluate comparative tissue responses. Furthermore, we introduce a custom composite loss function to address severe class imbalances and scoring inconsistencies across the five Swede score components. The proposed framework achieved 71.85\% accuracy and an 86.23\% AUC-ROC for three-class CIN grading, outperforming existing methods. For Swede score component prediction, the architecture achieved AUC-ROC values ranging from 75.7\% to 88.4\%, with the composite loss function yielding consistent F1-score improvements. Finally, the total predicted Swede Score, which ranges between 0 and 10, shows a Mean Absolute Error (MAE) of 1.489. The results show that the proposed method can pave the way towards developing AI-assisted colposcopy screening tools to support risk-based triage in resource-limited healthcare settings.
The dataset and source code are publicly available\footnote{\url{https://github.com/mHealthBuet/BUET-colposcopy}}.
\end{abstract}

\begin{IEEEkeywords}
Cervical cancer screening, colposcopy, cross-attention, deep learning, Swede score prediction.
\end{IEEEkeywords}
\section{Introduction}
\label{sec:introduction}
Cervical cancer is the fifth most common malignancy among women worldwide, with an estimated 604,000 new cases and 280,000 deaths in 2024~\cite{who_cervical_cancer}. The burden falls disproportionately on low- and middle-income countries (LMICs). These countries account for the large majority of cervical cancer deaths due to limited access to screening, vaccination, and treatment infrastructure. In Bangladesh, cervical cancer is the second most common cancer among women, with an estimated 8,068 new cases and 5,214 deaths being reported in 2018~\cite{uddin2023cervical}. Without intervention, projections suggest several hundred thousand additional preventable deaths in LMICs over the coming decades, underscoring the urgency of scalable screening solutions in resource-limited settings.

Cytology-based Papanicolaou (Pap) screening and human papillomavirus (HPV) vaccination have substantially reduced cervical cancer incidence in high-income countries. However, due to the lack of availability of Pap and HPV screening in LMICs,
colposcopy is used as the main screening and biopsy-guidance tool following an abnormal primary test. However, colposcopic interpretation is inherently subjective. Substantial inter- and intra-observer variability has been documented even among trained colposcopists. 

In LMICs, a critical shortage of trained colposcopists, combined with heavy clinical workloads and inconsistent documentation, prevents reliable triage even where screening coverage exists. Deep learning-assisted screening offers a path toward standardizing colposcopic feature recognition and extending expert-level diagnostic support to frontline health workers. However, there is a scarcity of well-annotated, multi-source colposcopy datasets representative of the LMIC population.

Prior deep learning approaches to colposcopy image analysis have largely treated CIN classification as a single-dataset, single-institution problem, with limited attention to generalizability across imaging equipment and patient populations (see Sec.~\ref{sec:related}). Moreover, no prior work has targeted automated and interpretable prediction of the Swede score~\cite{swede_original}. It is a five-component, clinically standardized colposcopic scoring rubric, including aceto-white uptake, margin and surface, vessel pattern, lesion size, and iodine staining, widely used to structure colposcopic assessment and guide referral decisions~\cite{ifcpc_nomenclature}. Existing fusion architectures for multimodal colposcopy images also typically rely on late concatenation of features extracted independently from each imaging modality.

In this work, we propose a dual cross-attention deep learning framework for automated CIN grading and Swede score component prediction. A novel dataset is also developed and annotated. The key contributions of this work are as follows:

\begin{itemize}
    \item The BUET Multi-Center Colposcopy Dataset, a novel dataset collected across multiple healthcare facilities in Dhaka, Bangladesh. It is the first colposcopy dataset of the country with transformation zone and Swede score annotations. In addition, 
    we also curated the IARC Colposcopy Image Bank dataset.
    \item A dual-stream cross-attention-based deep learning architecture for three-class CIN grading (Normal, CIN1, High Grade) that fuses paired cervigram features via a swapped cross-attention mechanism.
    \item The first end-to-end framework for predicting individual Swede score components from colposcopy images using a multi-head dual cross-attention network.
    \item A novel composite loss function to address class imbalance and also enhance the coherence of cross-component scoring of the Swede score components. The loss function combines weighted focal loss with effective-number-of-samples class weighting, a Huber-based total-score consistency term, and a head-correction mechanism. 
    
\end{itemize}

\section{Related Work}
\label{sec:related}

\subsection{CIN Classification}
Early work treated colposcopy classification as a conventional image classification problem. Saini~\textit{et al.}~\cite{saini_colponet} proposed ColpoNet, a DenseNet-inspired architecture for binary risk stratification (low-risk versus high-risk), achieving 81.35\% accuracy on the National Cancer Institute (NCI) colposcopy dataset and outperforming AlexNet, VGG16, ResNet50, and GoogLeNet baselines, though the approach relied on single static frames without region-of-interest segmentation. 

Recognizing that static frames discard the dynamic tissue response to contrast reagents, subsequent work combined spatial CNN feature extraction with recurrent sequence modeling. Yue~\textit{et al.}~\cite{yue_crcnn} developed a cervigram-based recurrent CNN (C-RCNN) combining an AlexNet-derived spatial encoder with an LSTM module and a multistate-aware fusion layer across saline, acetic acid, and iodine states, evaluated on 4,753 cervigrams from 679 cases, achieving 96.13\% accuracy and AUC values exceeding 0.94 across four classes. Chen~\textit{et al.}~\cite{chen_efficientnet_bigru} paired an EfficientNet-B0 backbone with a bidirectional GRU for cross-modal feature fusion on a larger cohort of 6,002 cases (18,006 images), reporting an ablation in which the recurrent fusion layer improved three-class accuracy from 88.32\% (backbone alone) to 91.18\%, demonstrating the value of explicitly modeling reagent-state transitions.

More recent work has incorporated self-attention to capture global context alongside local convolutional features. Mohammed~\textit{et al.}~\cite{mohammed_swin_cnn} combined a Swin Transformer stream with a CNN stream for binary classification on 898 IARC images, achieving 94\% AUC under five-fold cross-validation but a marked drop to 80--82\% AUC on independent test sets, illustrating the persistent challenge of out-of-distribution generalization. A related study fine-tuned Inception-ResNet-v2 and ResNet-152 across both CIN and LAST histopathological grading systems with test-time augmentation, reporting 87.7\% need-to-biopsy classification accuracy~\cite{cin_last_comparison}. Al-Hejri~\textit{et al.}~\cite{alhejri_vit_cytology} proposed an ensemble-CNN-to-ViT pipeline in the cytological domain, reporting near-perfect accuracies on single-cell datasets with Grad-CAM-based localization.

System-level deployment-oriented architectures have also emerged. Kalbhor~\textit{et al.}~\cite{kalbhor_deepcervicancer} proposed DeepCerviCancer, fusing separate colposcopy and cytology sub-networks (DeepColpo, DeepCyto+) through conventional classifiers, evaluated on a small same-patient cohort. Skerrett~\textit{et al.}~\cite{skerrett_pocket_colposcope} targeted low-cost, field-deployable hardware using a Pocket Colposcope, combining a class-balanced weighted cross-entropy loss with parallel acetic-acid/green-light feature fusion to achieve 0.87 AUC with 75\% sensitivity and 88\% specificity. Fang~\textit{et al.}~\cite{fang_shufflenet} proposed a lightweight ShuffleNet with squeeze-and-excitation and selective-kernel attention for edge deployment, achieving 81.38\% five-class accuracy with substantially fewer parameters than VGG-16 or ResNet-34, though trained on a single institution's data.

Overall, most studies in this area considered training and validation on a single institutional dataset. Multimodal fusion is typically achieved through late concatenation rather than learned cross-attention, and class imbalance, when addressed, did not consider class-aware loss reweighting with task-specific consistency constraints.

\subsection{Swede Score Prediction}
Compared to CIN classification, automated Swede score prediction remains largely unexplored, constrained in part by the scarcity of annotated colposcopy datasets. Ren~\textit{et al.}~\cite{ren_annocerv} introduced AnnoCerv, a multimodal colposcopy dataset of 527 images from 100 clinical records combining acetic acid, iodine, and green-filtered modalities with expert-informed, color-coded morphological annotations, addressing the annotation scarcity problem. 
However, to the best of our knowledge, no peer-reviewed study has framed the Swede score component prediction using an end-to-end deep learning model. 
\section{Datasets}
\label{sec:dataset}

\begin{figure}[!t]
\centering
\subfloat[Normal saline image]{\includegraphics[width=0.48\linewidth]{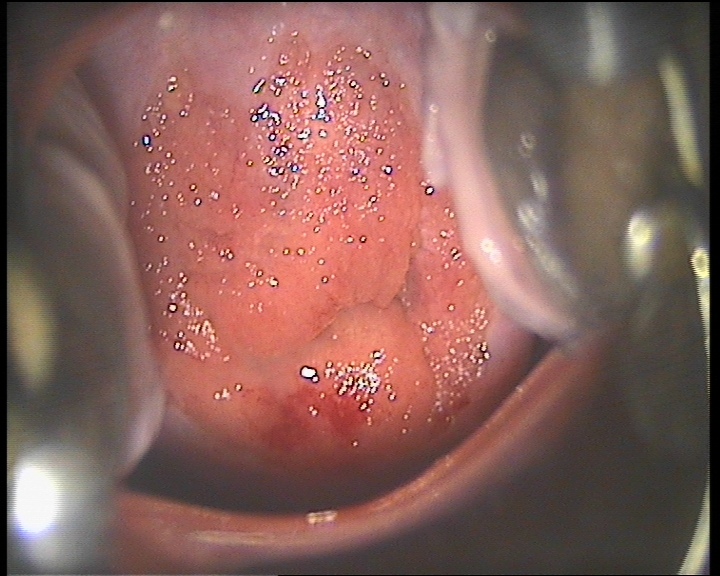}}
\hfill
\subfloat[Green filtered image]{\includegraphics[width=0.48\linewidth]{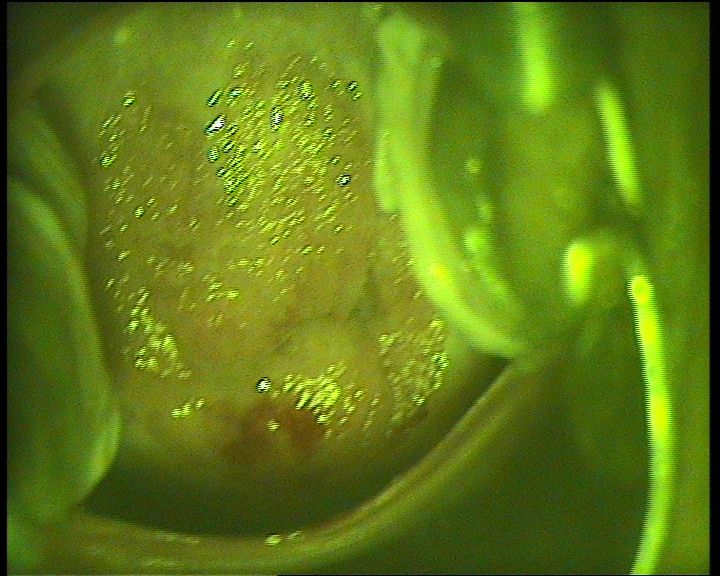}}

\vspace{0.5em}

\subfloat[Aceto uptake image]{\includegraphics[width=0.48\linewidth]{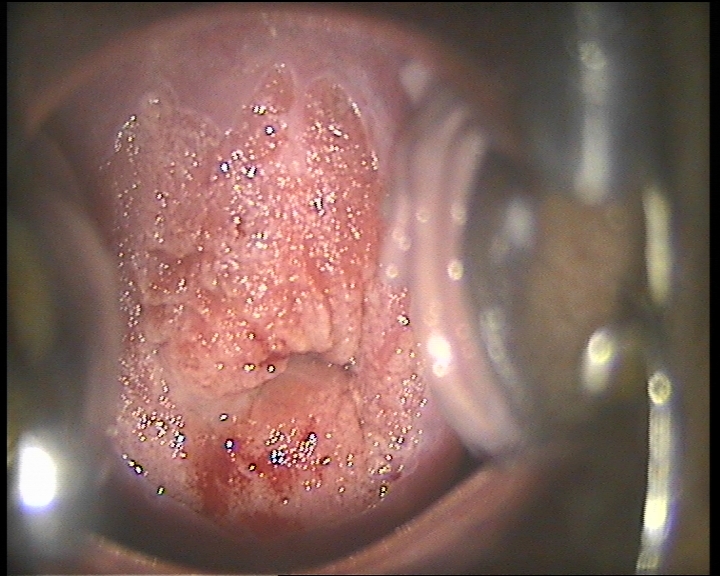}}
\hfill
\subfloat[Iodine uptake image]{\includegraphics[width=0.48\linewidth]{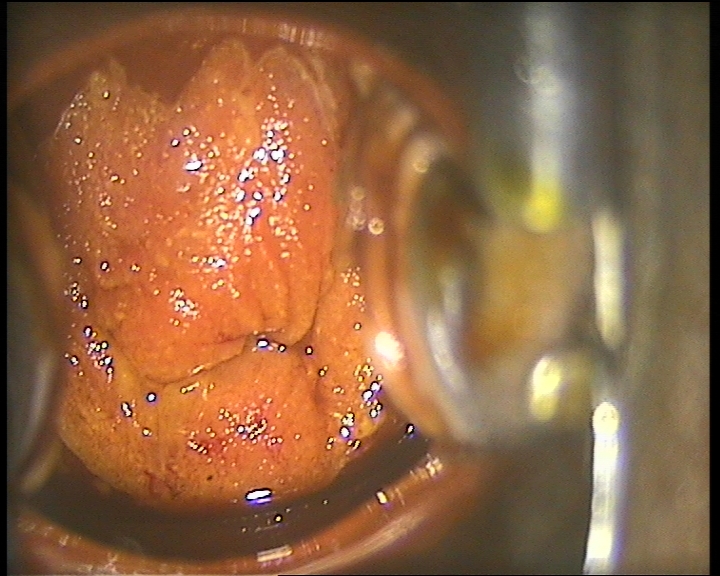}}

\caption{Four sample images from a single case in the dataset, shown under the four visualization modalities: (a) normal saline, (b) green filter, (c) aceto uptake, and (d) iodine uptake.}
\label{fig:dmch_samples}
\end{figure}

\subsection{IARC Colposcopy Image Bank}
The IARC Colposcopy Image Bank~\cite{iarc_image_bank} is designed to support the development, evaluation, and validation of machine learning algorithms for the early identification of cervical cancer and precancerous conditions. The subset used in this study comprises 756 images from 189 full case reports. Each case contained detailed clinical metadata, including HPV test status, transformation zone type, lesion location, and grade, and a fully structured Swede score. Images span four visualization states: normal saline, acetic acid, Lugol's iodine, and green-filter visualization. The Swede score distribution is substantially imbalanced, with normal and low-grade findings far outnumbering high-grade findings.

\subsection{BUET Multi-Center Colposcopy Dataset}
\label{subsec:buet_multicenter_dataset}

To address the limited representation of Bangladeshi populations in existing colposcopy literature, we curated a novel, multi-center dataset in collaboration with multiple hospitals across Dhaka, Bangladesh. Collecting multi-center data captures demographic diversity and imaging device variability, thereby improving model robustness to domain variability.

The data collection study was approved by the Ethical Review Committee (ERC) of the Department of Biomedical Engineering, BUET (Ref: BME/ERC/2026/03) and by the Institutional Review Boards (IRBs) of the collaborating hospitals. A strict anonymization process was maintained, and participants' right to withdraw was preserved throughout the informed consent process.

For each patient, a four-image series was obtained as described in Fig.~\ref{fig:dmch_samples}. Clinical metadata and labels included transformation zone type and Swede score, where the total Swede score can be mapped to a provisional clinical diagnosis. Selected CIN2 and CIN3+ cases were reviewed by a second expert colposcopist to validate case grading prior to inclusion. After filtering for cases with adequate Swede score documentation, the BUET Colposcopy dataset comprises 768 patient cases, yielding 3,072 images in total. Table~\ref{tab:buet_multicenter_swede_dist} summarizes the distribution of scores across the five components.

\begin{table}[!t]
\caption{Score Distribution of the Five Swede Score Characteristics in the BUET Multi-Center Colposcence Dataset}
\label{tab:buet_multicenter_swede_dist}
\centering
\begin{tabular}{lccc}
\hline
\textbf{Characteristic} & \textbf{Score 0} & \textbf{Score 1} & \textbf{Score 2} \\
\hline
Aceto-White Uptake & 293 & 409 & 66 \\
Margin and Surface & 441 & 312 & 15 \\
Vessels & 635 & 84 & 49 \\
Lesion Size & 336 & 357 & 75 \\
Iodine Staining & 207 & 439 & 122 \\
\hline
\end{tabular}
\end{table}

\subsection{Combined Dataset}
\label{subsec:dataset_combination}
The two datasets were combined for the model training and validation. Images of insufficient quality or without proper annotations were excluded. The combined dataset was harmonized into a three-class CIN grading scheme (Normal, CIN1, High Grade). High Grade merges the original CIN2 and CIN3 categories to better reflect clinically established high-grade lesion terminology and to mitigate the severe class sparsity of CIN3 alone. Table~\ref{tab:combined_cin_dist} presents the resulting combined CIN distribution. For Swede score prediction, the same two datasets were combined and harmonized into a five-component scoring scheme (Aceto Uptake, Iodine Uptake, Vessel Pattern, Margin, Lesion Size), each scored 0--2. Table~\ref{tab:combined_swede_dist} presents the resulting score distribution of total 957 cases with 3828 images. Cases with a missing label were eliminated during training.

\begin{table}[!t]
\caption{Score Distribution of the Five Swede Score Components in the Combined Dataset}
\label{tab:combined_swede_dist}
\centering
\begin{tabular}{lccc}
\hline
\textbf{Characteristic} & \textbf{Score 0} & \textbf{Score 1} & \textbf{Score 2} \\
\hline
Aceto Uptake & 380 & 444 & 133 \\
Iodine Uptake & 239 & 507 & 168 \\
Vessel Pattern & 685 & 189 & 83 \\
Margin & 530 & 341 & 86 \\
Lesion Size & 422 & 400 & 135 \\
\hline
\end{tabular}
\end{table}

\begin{table}[!t]
\caption{Frequency Distribution of CIN Classification Data in the Combined Dataset}
\label{tab:combined_cin_dist}
\centering
\begin{tabular}{lc}
\hline
\textbf{Diagnosis} & \textbf{Frequency} \\
\hline
Normal & 207 \\
CIN1 & 492 \\
High Grade & 258 \\
\hline
\end{tabular}
\end{table}

\definecolor{pFill}{HTML}{7F77DD}   \definecolor{pStroke}{HTML}{534AB7}
\definecolor{pLight}{HTML}{AFA9EC}  \definecolor{pText}{HTML}{26215C}
\definecolor{pBox}{HTML}{EEEDFE}
\definecolor{tFill}{HTML}{1D9E75}   \definecolor{tStroke}{HTML}{0F6E56}
\definecolor{tLight}{HTML}{5DCAA5}  \definecolor{tText}{HTML}{04342C}
\definecolor{tBox}{HTML}{E1F5EE}
\definecolor{coralFill}{HTML}{FAECE7} \definecolor{coralStroke}{HTML}{993C1D}
\definecolor{coralText}{HTML}{4A1B0C}
\definecolor{amberFill}{HTML}{FAEEDA} \definecolor{amberStroke}{HTML}{854F0B}
\definecolor{amberText}{HTML}{412402}
\definecolor{cbamFill}{HTML}{F7F5F0}  \definecolor{grayStroke}{HTML}{5F5E5A}
\definecolor{grayTitle}{HTML}{2C2C2A}
\definecolor{chFill}{HTML}{EAF3DE}    \definecolor{chStroke}{HTML}{3B6D11}
\definecolor{chText}{HTML}{173404}
\definecolor{poolFill}{HTML}{F1EFE8}
\definecolor{fcBlue}{HTML}{378ADD}    \definecolor{fcBlueS}{HTML}{185FA5}
\definecolor{fcGreen}{HTML}{639922}   \definecolor{fcGreenS}{HTML}{3B6D11}
\definecolor{outFront}{HTML}{D85A30}  \definecolor{outTop}{HTML}{F0997B}
\definecolor{arrGray}{HTML}{4A4A4A}   \definecolor{thinGray}{HTML}{8A8A8A}

\newcommand{\cuboid}[8]{%
  \fill[#6,fill opacity=0.5]  (#1,#2) -- (#1+12,#2-12) -- (#1+#3+12,#2-12) -- (#1+#3,#2) -- cycle;
  \fill[#7,fill opacity=0.35] (#1+#3,#2) -- (#1+#3+12,#2-12) -- (#1+#3+12,#2+#4-12) -- (#1+#3,#2+#4) -- cycle;
  \fill[#5,fill opacity=0.24] (#1,#2) rectangle (#1+#3,#2+#4);
  \draw[#8,line width=0.7pt] (#1,#2) rectangle (#1+#3,#2+#4);
  \draw[#8,line width=0.7pt] (#1,#2) -- (#1+12,#2-12) -- (#1+#3+12,#2-12) -- (#1+#3,#2);
  \draw[#8,line width=0.7pt] (#1+#3,#2) -- (#1+#3+12,#2-12) -- (#1+#3+12,#2+#4-12) -- (#1+#3,#2+#4);
}

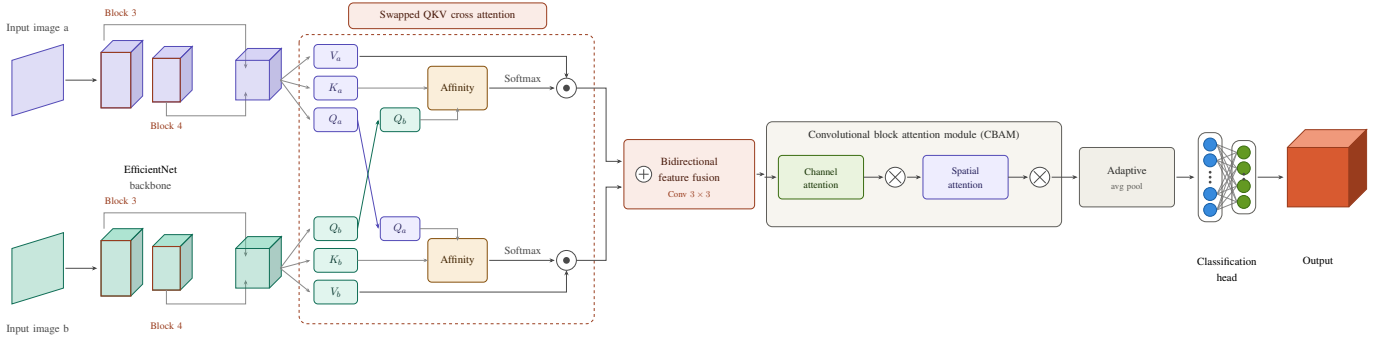
\begin{figure*}[t]
\centering
\resizebox{\textwidth}{!}{
\begin{tikzpicture}[x=1pt,y=-1pt,
    arr/.style={draw=arrGray,line width=1pt,-{Stealth[length=5pt,width=4pt]}},
    thn/.style={draw=thinGray,line width=0.7pt,-{Stealth[length=4pt,width=3pt]}},
    lbl/.style={font=\fontsize{11}{12}\selectfont,text=arrGray},
    box/.style={rounded corners=4pt,line width=0.8pt},
    font=\sffamily]

\node[lbl] at (66,88) {Input image a};
\fill[pFill,fill opacity=0.25] (40,120)--(92,104)--(92,160)--(40,176)--cycle;
\draw[pStroke,line width=0.7pt] (40,120)--(92,104)--(92,160)--(40,176)--cycle;
\draw[arr] (94,140)--(126,140);
\cuboid{130}{112}{30}{56}{pFill}{pLight}{pStroke}{pStroke}
\cuboid{182}{118}{28}{44}{pFill}{pLight}{pStroke}{pStroke}
\draw[coralStroke,line width=1.1pt] (130,112) rectangle (160,168);
\draw[coralStroke,line width=1.1pt] (182,118) rectangle (210,162);
\node[coralStroke,font=\fontsize{10}{11}\selectfont] at (150,72) {Block 3};
\node[coralStroke,font=\fontsize{10}{11}\selectfont] at (196,186) {Block 4};
\draw[thn] (133,100)--(133,84)--(276,84)--(276,128);
\draw[thn] (196,162)--(196,176)--(276,176)--(276,152);
\cuboid{266}{120}{34}{40}{pFill}{pLight}{pStroke}{pStroke}

\node[lbl] at (66,392) {Input image b};
\fill[tFill,fill opacity=0.25] (40,310)--(92,294)--(92,350)--(40,366)--cycle;
\draw[tStroke,line width=0.7pt] (40,310)--(92,294)--(92,350)--(40,366)--cycle;
\draw[arr] (94,330)--(126,330);
\cuboid{130}{302}{30}{56}{tFill}{tLight}{tStroke}{tStroke}
\cuboid{182}{308}{28}{44}{tFill}{tLight}{tStroke}{tStroke}
\draw[coralStroke,line width=1.1pt] (130,302) rectangle (160,358);
\draw[coralStroke,line width=1.1pt] (182,308) rectangle (210,352);
\node[coralStroke,font=\fontsize{10}{11}\selectfont] at (150,262) {Block 3};
\node[coralStroke,font=\fontsize{10}{11}\selectfont] at (196,388) {Block 4};
\draw[thn] (133,290)--(133,274)--(276,274)--(276,318);
\draw[thn] (196,352)--(196,366)--(276,366)--(276,342);
\cuboid{266}{310}{34}{40}{tFill}{tLight}{tStroke}{tStroke}

\node[font=\fontsize{11}{12}\selectfont,text=black] at (180,230) {EfficientNet};
\node[lbl] at (180,246) {backbone};

\draw[coralStroke,line width=0.8pt,dashed,rounded corners=8pt] (330,94) rectangle (628,386);
\draw[coralStroke,line width=0.8pt,fill=coralFill,rounded corners=5pt] (380,62) rectangle (580,88);
\node[coralText,font=\fontsize{11}{12}\selectfont] at (480,75) {Swapped QKV cross attention};

\draw[thn] (311,140)--(341,116);
\draw[thn] (311,140)--(341,148);
\draw[thn] (311,140)--(341,180);
\foreach \yy/\lab in {104/{$V_a$},136/{$K_a$},168/{$Q_a$}}{
  \draw[box,fill=pBox,draw=pStroke] (345,\yy) rectangle (389,\yy+24);
  \node[pText,font=\fontsize{11}{12}\selectfont] at (367,\yy+12) {\lab};}

\draw[thn] (311,330)--(341,290);
\draw[thn] (311,330)--(341,322);
\draw[thn] (311,330)--(341,354);
\foreach \yy/\lab in {278/{$Q_b$},310/{$K_b$},342/{$V_b$}}{
  \draw[box,fill=tBox,draw=tStroke] (345,\yy) rectangle (389,\yy+24);
  \node[tText,font=\fontsize{11}{12}\selectfont] at (367,\yy+12) {\lab};}

\draw[pStroke,line width=1pt,-{Stealth[length=5pt]}] (389,180)--(410,290);
\draw[tStroke,line width=1pt,-{Stealth[length=5pt]}] (389,290)--(410,180);
\draw[box,fill=tBox,draw=tStroke] (412,168) rectangle (452,192);
\node[tText,font=\fontsize{11}{12}\selectfont] at (432,180) {$Q_b$};
\draw[box,fill=pBox,draw=pStroke] (412,278) rectangle (452,302);
\node[pText,font=\fontsize{11}{12}\selectfont] at (432,290) {$Q_a$};

\draw[box,fill=amberFill,draw=amberStroke] (460,126) rectangle (520,170);
\node[amberText,font=\fontsize{11}{12}\selectfont] at (490,148) {Affinity};
\draw[box,fill=amberFill,draw=amberStroke] (460,300) rectangle (520,344);
\node[amberText,font=\fontsize{11}{12}\selectfont] at (490,322) {Affinity};

\draw[thn] (389,148)--(460,148);

\draw[thn] (452,180)--(490,180)--(490,168);

\draw[thn] (389,322)--(460,322);

\draw[thn] (452,290)--(490,290)--(490,302);

\draw[arr] (391,116)--(600,116)--(600,138);
\draw[arr] (391,354)--(600,354)--(600,332);
\draw[arr] (522,148)--(588,148);
\draw[arr] (522,322)--(588,322);
\node[lbl] at (556,138) {Softmax};
\node[lbl] at (556,312) {Softmax};
\draw[fill=white,draw=arrGray,line width=0.9pt] (600,148) circle (10);
\fill[arrGray] (600,148) circle (3.4);
\draw[fill=white,draw=arrGray,line width=0.9pt] (600,322) circle (10);
\fill[arrGray] (600,322) circle (3.4);

\draw[arr] (610,148)--(640,148)--(640,222)--(656,222);
\draw[arr] (610,322)--(640,322)--(640,248)--(656,248);
\draw[box,fill=coralFill,draw=coralStroke] (658,200) rectangle (790,270);
\draw[fill=white,draw=arrGray,line width=0.8pt] (678,235) circle (8);
\draw[arrGray,line width=0.8pt] (673,235)--(683,235);
\draw[arrGray,line width=0.8pt] (678,230)--(678,240);
\node[coralText,font=\fontsize{11}{12}\selectfont] at (724,222) {Bidirectional};
\node[coralText,font=\fontsize{11}{12}\selectfont] at (724,238) {feature fusion};
\node[coralStroke,font=\fontsize{9}{10}\selectfont] at (724,255) {Conv $3\times3$};
\draw[arr] (792,235)--(800,235);

\draw[box,fill=cbamFill,draw=grayStroke,rounded corners=6pt] (802,182) rectangle (1100,288);
\node[grayTitle,font=\fontsize{11}{12}\selectfont] at (951,196) {Convolutional block attention module (CBAM)};
\draw[box,fill=chFill,draw=chStroke,line width=1pt] (814,216) rectangle (900,260);
\node[chText,font=\fontsize{10}{11}\selectfont] at (857,232) {Channel};
\node[chText,font=\fontsize{10}{11}\selectfont] at (857,246) {attention};
\draw[box,fill=pBox,draw=pStroke,line width=1pt] (960,216) rectangle (1046,260);
\node[pText,font=\fontsize{10}{11}\selectfont] at (1003,232) {Spatial};
\node[pText,font=\fontsize{10}{11}\selectfont] at (1003,246) {attention};

\foreach \cx in {932,1078}{
  \draw[fill=white,draw=arrGray,line width=0.9pt] (\cx,238) circle (10);
  \draw[arrGray,line width=1pt] (\cx-6,232)--(\cx+6,244);
  \draw[arrGray,line width=1pt] (\cx+6,232)--(\cx-6,244);}
\draw[arr] (800,238)--(812,238);
\draw[arr] (900,238)--(920,238);
\draw[arr] (944,238)--(958,238);
\draw[arr] (1046,238)--(1066,238);
\draw[arr] (1090,238)--(1116,238);

\draw[box,fill=poolFill,draw=grayStroke] (1118,208) rectangle (1214,268);
\node[grayTitle,font=\fontsize{11}{12}\selectfont] at (1166,232) {Adaptive};
\node[grayStroke,font=\fontsize{9}{10}\selectfont] at (1166,248) {avg pool};
\draw[arr] (1216,238)--(1234,238);

\foreach \sy in {205,221,255,271}{
  \foreach \ty in {213,229,247,263}{
    \draw[thinGray,line width=0.35pt] (1257,\sy)--(1277,\ty);}}
\draw[grayStroke,line width=0.8pt,rounded corners=8pt] (1238,193) rectangle (1262,283);
\draw[grayStroke,line width=0.8pt,rounded corners=8pt] (1272,201) rectangle (1296,275);
\foreach \cy in {205,221,255,271}{\draw[fill=fcBlue,draw=fcBlueS,line width=0.7pt] (1250,\cy) circle (6.5);}
\foreach \cy in {213,229,247,263}{\draw[fill=fcGreen,draw=fcGreenS,line width=0.7pt] (1284,\cy) circle (6.5);}
\foreach \cy in {233,238,243}{\fill[arrGray] (1250,\cy) circle (1.4);}
\foreach \cy in {235,240}{\fill[arrGray] (1284,\cy) circle (1.4);}
\node[font=\fontsize{11}{12}\selectfont,text=black] at (1267,323) {Classification};
\node[font=\fontsize{11}{12}\selectfont,text=black] at (1267,339) {head};
\draw[arr] (1298,238)--(1324,238);

\fill[outTop] (1328,208)--(1346,190)--(1408,190)--(1390,208)--cycle;
\fill[coralStroke] (1390,208)--(1408,190)--(1408,250)--(1390,268)--cycle;
\fill[outFront] (1328,208) rectangle (1390,268);
\draw[coralStroke,line width=0.8pt] (1328,208) rectangle (1390,268);
\draw[coralStroke,line width=0.8pt] (1328,208)--(1346,190)--(1408,190)--(1390,208);
\draw[coralStroke,line width=0.8pt] (1390,208)--(1408,190)--(1408,250)--(1390,268);
\node[font=\fontsize{11}{12}\selectfont,text=black] at (1359,323) {Output};

\end{tikzpicture}
}
\caption{Architecture of the proposed dual-stream cross-attention network. Paired multimodal cervigrams are encoded via parallel EfficientNet backbones and fused using a swapped QKV cross-attention module.}
\label{fig:model1_arch}
\end{figure*}

\section{Methodology}
\label{sec:methodology}

\subsection{Overall Framework}
Our framework comprises two streams trained on the combined dataset (Sec.~\ref{sec:dataset}-C): (i) a three-class CIN classification model (Normal, CIN1, High Grade), and (ii) a multi-head Swede score component prediction model whose five outputs are summed to produce a total Swede score in the range 0--10.

\subsection{Data Annotation}
As routine colposcopy reports frequently lack a component-level breakdown of the Swede score, retrospective annotation was required. Experienced colposcopist from each participating hospital systematically reviewed the clinical cases and manually annotated the individual Swede score components.

\definecolor{inFill}{HTML}{EEEDFE}   \definecolor{inStroke}{HTML}{534AB7}
\definecolor{modFill}{HTML}{FAEEDA}  \definecolor{modStroke}{HTML}{854F0B}
\definecolor{predFill}{HTML}{E1F5EE} \definecolor{predStroke}{HTML}{0F6E56}
\definecolor{totFill}{HTML}{F7F5F0}  \definecolor{totStroke}{HTML}{2C2C2A}

\begin{figure*}[t]
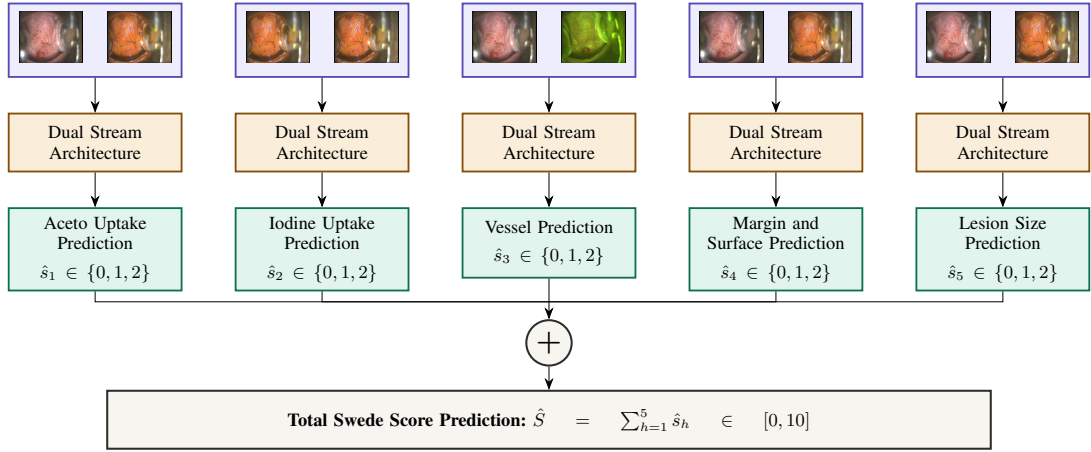

\centering
\resizebox{0.8\textwidth}{!}{
\begin{tikzpicture}[
    inbox/.style={rectangle, draw=inStroke, fill=inFill, line width=0.9pt, align=center, minimum height=1cm, text width=2.7cm, font=\small, inner sep=4pt},
    modbox/.style={rectangle, draw=modStroke, fill=modFill, line width=0.9pt, align=center, minimum height=1cm, text width=2.7cm, font=\small, inner sep=4pt},
    predbox/.style={rectangle, draw=predStroke, fill=predFill, line width=0.9pt, align=center, minimum height=1.2cm, text width=2.7cm, font=\small, inner sep=4pt},
    sumbox/.style={rectangle, draw=totStroke, fill=totFill, line width=1pt, align=center, minimum height=1cm, text width=15cm, font=\small},
    concat/.style={circle, draw=totStroke, fill=totFill, line width=1pt, inner sep=2pt, font=\Large},
    arr/.style={-{Stealth[length=2mm]}}
]

\node[inbox] (in1) {
\begin{tabular}{@{}cc@{}}
\includegraphics[width=1.1cm]{Sample_image/003.jpg} & \includegraphics[width=1.1cm]{Sample_image/004.jpg} \\
\end{tabular}};
\node[modbox, below=0.6cm of in1] (m1) {Dual Stream Architecture};
\node[predbox, below=0.6cm of m1] (p1) {Aceto Uptake Prediction\\[1ex] $\hat{s}_1 \in \{0,1,2\}$};

\node[inbox, right=0.9cm of in1] (in2) {
\begin{tabular}{@{}cc@{}}
\includegraphics[width=1.1cm]{Sample_image/004.jpg} & \includegraphics[width=1.1cm]{Sample_image/004.jpg} \\
\end{tabular}};
\node[modbox, below=0.6cm of in2] (m2) {Dual Stream Architecture};
\node[predbox, below=0.6cm of m2] (p2) {Iodine Uptake Prediction\\[1ex] $\hat{s}_2 \in \{0,1,2\}$};

\node[inbox, right=0.9cm of in2] (in3) {
\begin{tabular}{@{}cc@{}}
\includegraphics[width=1.1cm]{Sample_image/003.jpg} & \includegraphics[width=1.1cm]{Sample_image/002.jpg} \\
\end{tabular}};
\node[modbox, below=0.6cm of in3] (m3) {Dual Stream Architecture};
\node[predbox, below=0.6cm of m3] (p3) {Vessel Prediction\\[1ex] $\hat{s}_3 \in \{0,1,2\}$};

\node[inbox, right=0.9cm of in3] (in4) {%
\begin{tabular}{@{}cc@{}}
\includegraphics[width=1.1cm]{Sample_image/003.jpg} & \includegraphics[width=1.1cm]{Sample_image/004.jpg} \\
\end{tabular}};
\node[modbox, below=0.6cm of in4] (m4) {Dual Stream Architecture};
\node[predbox, below=0.6cm of m4] (p4) {Margin and Surface Prediction\\[1ex] $\hat{s}_4 \in \{0,1,2\}$};

\node[inbox, right=0.9cm of in4] (in5) {%
\begin{tabular}{@{}cc@{}}
\includegraphics[width=1.1cm]{Sample_image/003.jpg} & \includegraphics[width=1.1cm]{Sample_image/004.jpg} \\
\end{tabular}};
\node[modbox, below=0.6cm of in5] (m5) {Dual Stream Architecture};
\node[predbox, below=0.6cm of m5] (p5) {Lesion Size Prediction\\[1ex] $\hat{s}_5 \in \{0,1,2\}$};

\foreach \i in {1,...,5} {
  \draw[arr] (in\i) -- (m\i);
  \draw[arr] (m\i) -- (p\i);
}

\coordinate (merge) at ($(p3.south) + (0,-0.4)$);
\node[concat, below=0.3cm of merge] (plus) {$\boldsymbol{+}$};
\node[sumbox, below=0.4cm of plus] (total) {\textbf{Total Swede Score Prediction:} $\hat{S} = \sum_{h=1}^{5} \hat{s}_h \in [0,10]$};

\foreach \i in {1,...,5} {
  \draw[-] (p\i.south) -- (p\i.south |- merge) -- (merge);
}

\draw[arr] (merge) -- (plus.north);
\draw[arr] (plus.south) -- (total.north);

\end{tikzpicture}
}
\caption{Overall architecture of the proposed multi-head Swede score prediction model. Five task-specific branches each receive a dedicated image pair, pass through the proposed dual stream classification architecture, and output an individual characteristic score $\hat{s}_h \in \{0,1,2\}$. The total Swede score $\hat{S} = \sum_{h=1}^{5} \hat{s}_h \in [0,10]$ is obtained by aggregating all five predicted scores.}
\label{fig:swede_architecture}
\end{figure*}
\subsection{Data Preprocessing}
\subsubsection{Segmentation}
 Manual region-of-interest (ROI) segmentation was performed using the Image Processing Toolbox in MATLAB across three imaging modalities per case (acetic acid, iodine, and green-filtered), using a freehand tool to delineate cervix boundaries and remove extraneous background and instrument artifacts.
\subsubsection{Augmentation}
For CIN classification, paired acetic acid and iodine images were augmented identically per sample to preserve pixel-level anatomical correspondence: random horizontal flip ($p=0.5$), random vertical flip ($p=0.5$), and random rotation sampled uniformly from $[-20^{\circ}, +20^{\circ}]$, followed by resizing to $224\times224$ and ImageNet channel-wise normalization. No photometric distortion was applied to avoid corrupting diagnostically relevant aceto-white and iodine color responses. 
For the Swede score prediction, the acetic acid, iodine, and green-filtered images for each sample were stacked into a unified 9-channel image during training. It guaranteed identical geometric augmentation (random flip $p=0.5$, rotation $\pm20^{\circ}$, zoom $\pm10\%$, contrast jitter $\pm5\%$) across all three modalities before being split back into 3-channel inputs.
\subsubsection{ROI cropping}
For the Swede score prediction, a deterministic background-cropping was performed using OpenCV. Non-informative black background was removed via grayscale intensity thresholding ($\tau=15$), that tightly bound the cervical tissue region. 
\subsection{Proposed Architecture}
Fig. \ref{fig:model1_arch} illustrates the proposed dual-stream architecture. The model is designed to emulate the diagnostic workflow of a colposcopist during both CIN classification and Swede score prediction. The framework simultaneously extracts and integrates complementary features from two distinct modalities.

\subsubsection{Swapped Query-Key-Value Cross-Attention}
A swapped QKV cross-attention mechanism is applied to model cross-modal diagnostic correspondence. This computes the Query from the acetic acid stream and matches it against the Key and Value of the iodine stream. Furthermore, it symmetrically computes the Query from the iodine stream against the acetic acid Key and Value.

 \begin{align}
    \text{Attn}_{a \to b} &= \text{softmax}\left(\frac{Q_a K_b^\top}{\sqrt{d_k}}\right)V_b \label{eq:attn_a2i} \\
    \text{Attn}_{b \to a} &= \text{softmax}\left(\frac{Q_b K_a^\top}{\sqrt{d_k}}\right)V_a \label{eq:attn_i2a}
\end{align}

where $Q_a$, $K_a$, $V_a$ and $Q_b$, $K_b$, $V_b$ are the Query, Key, and Value projections of the input image a and input image b streams, respectively, and $d_k$ represents the scaling factor (dimension of the keys). This allows the network to identify spatial regions where both modalities jointly indicate a pathological change, rather than processing each modality independently before late fusion.

\subsubsection{Bidirectional Feature Fusion}
 \label{subsubsec:dcaf}
The two cross-attended 256-channel feature maps are fused via a $3\times3$ convolutional Bidirectional Feature Fusion layer into a single joint representation. This step synthesizes the independently attended acetic acid and iodine representations. Consequently, these are turned into a single multidimensional tensor.

\begin{equation}
    F_{\text{fused}} = \text{Conv}_{3\times3}(\text{Attn}_{a \to b} + \text{Attn}_{b \to a})
    \label{eq:fused_representation}
\end{equation}

\subsubsection{Convolutional Block Attention Module}
This representation is then refined by a Convolutional Block Attention Module (CBAM)~\cite{cbam_woo}. The module sequentially applies channel attention via combined global average and max pooling. Additionally, spatial attention is applied via a $7\times7$ convolution over channel-pooled feature maps. The process suppresses background artifacts such as specular glare. Concurrently, it emphasizes diagnostically relevant lesion boundaries and vascular patterns. Finally, a three-class classification is performed. 

\subsection{Proposed CIN Classification Model}
\label{subsec:model1}
The proposed CIN classification model architecture (Fig.~\ref{fig:model1_arch}) processes paired acetic acid and iodine cervigrams. Each stream uses an ImageNet-pretrained EfficientNet-B4 backbone. This pre-trained backbone extracts intermediate feature maps at Blocks 3 and 4. A feature fusion module is included within each stream, to ensure that the model analyzes the general morphology of the cervix and microscopic cellular anomalies simultaneously. The fused representations are subsequently processed with a Convolutional Block Attention Module (CBAM). Finally, the multi-modal tensor is compressed via global adaptive average pooling and passed through a fully connected classifier network. Dropout is used as regularization. At the end, a final three-class diagnostic probability is generated.

\begin{figure}[!t]
\centering
\includegraphics[width=\linewidth]{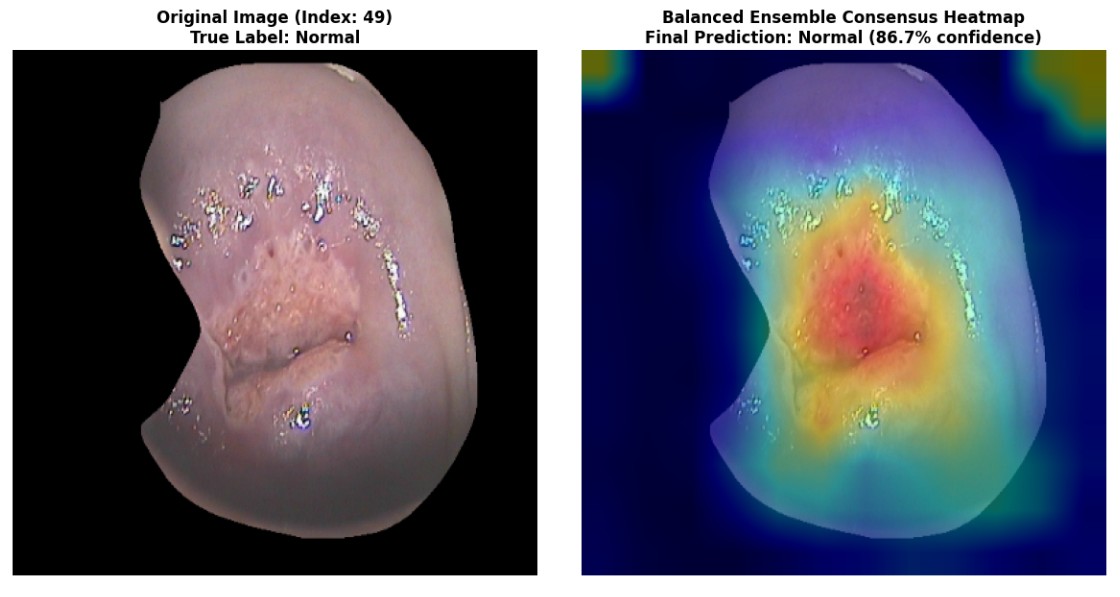}
\caption{Heatmap visualization using LayerCAM on a colposcopic image.}
\end{figure}

Patient-level data leakage was prevented using a two-stage splitting protocol: GroupShuffleSplit reserved 20\% of patients as a held-out test set. The remaining 80\% were split via five-fold StratifiedGroupKFold cross-validation. Patient identifier is enforced as the group variable in both stages.

\subsection{Proposed Swede Score Prediction Model}
We extend the dual-stream cross-attention architecture into a multi-input, multi-output, multi-task framework for the Swede score prediction task. It comprises five independent prediction heads, one for each Swede score component.  
Each component is formulated as an independent three-class classification problem ($s_h \in \{0,1,2\}$) simultaneously, and the component scores are summed to obtain the total predicted Swede score $\hat{S} = \sum_{h=1}^{5}\hat{s}_h \in [0,10]$ (Fig.~\ref{fig:swede_architecture}).

As shown in Fig.~\ref{fig:swede_architecture}, each head receives a dedicated pair of colposcopic modalities chosen to reflect how an expert colposcopist visually cross-references different views while scoring. Each head instantiates the dual-stream cross-attention architecture. Two EfficientNet backbones extract features that are fused via Dual Cross-Attention Fusion (Section~\ref{subsubsec:dcaf}) and refined with CBAM before a shared classification head. The backbone pairings in Table~\ref{tab:backbone_assignment} were finalized empirically. 

As the Swede score components are predicted jointly from the same case and each is imbalanced differently, we adopted iterative multilabel stratification~\cite{sechidis_multilabel2011} to balance the combined label distribution across folds. A stratified 80:20 split reserves a held-out test set. The remaining cases are partitioned into three stratified folds for cross-validation, with both steps balancing the joint five-label distribution. As each row corresponds to a single patient case, this stratification is performed directly at the case level.

We also implemented a component-specific early stopping and weight-freezing method. Because five individual scores are trained at a time, the optimal number of epochs varies across scores. Once a specific component reaches the plateau region of validation loss, we freeze the weights of that component while the training continues until all branches have fully converged. This approach reduces underfitting or overfitting of any single branch.

\begin{table}[!t]
\caption{EfficientNet Backbone Pair Assignments per Prediction Head}
\label{tab:backbone_assignment}
\centering

\resizebox{\columnwidth}{!}{%
\begin{tabular}{lll}
\hline
\textbf{Head} & \textbf{Backbone A} & \textbf{Backbone B} \\
\hline
Aceto-White Uptake & EfficientNetB4 (Aceto) & EfficientNetB0 (Iodine) \\
Margin and Surface & EfficientNetB0 (Aceto) & EfficientNetB0 (Iodine) \\
Vessel Pattern & EfficientNetB4 (Aceto) & EfficientNetB4 (Green Filter) \\
Lesion Size & EfficientNetB0 (Aceto) & EfficientNetB0 (Iodine) \\
Iodine Staining & EfficientNetB0 (Iodine) & EfficientNetB4 (Iodine) \\
\hline
\end{tabular}%
}
\end{table}

\subsection{Novel Composite Loss Function}
\label{subsec:loss}
The proposed loss function addresses the severe per-head class imbalance and ensures the consistency of the aggregate Swede score across the five independently trained heads.

\subsubsection{Weighted Focal Loss}
Each head is supervised with Focal Loss ($\gamma=2.0$). It down-weights the contribution of easily classified samples and amplifies the gradient signal from hard, typically minority-class:
\begin{equation}
\mathcal{L}_{\text{focal}} = -\sum_{h}(1-p_t^h)^{\gamma}\log(p_t^h)
\label{eq:focal}
\end{equation}
Per-head class weights are computed using the Effective Number of Samples (ENS) formulation~\cite{cui_ens_2019}. It accounts for information redundancy between samples. For \textit{n} number of samples in the class:
\begin{equation}
E_n = \frac{1-\beta^n}{1-\beta}
\label{eq:ens}
\end{equation}
where $\beta \in [0,1)$ is tuned per head according to class imbalance severity. The class weight $w = 1/E_n$ is normalized so that the weights sum to the number of classes.

\subsubsection{Total Score Huber Loss}
Since small per-head errors can aggregate if multiple heads err in the same direction, a Huber loss~\cite{huber_1964} is applied between $\hat{S}=\sum_{h=1}^{5}\hat{s}_h$ and $S$ to ensure the aggregate predicted score $\hat{S}$ matches the true total score $S$ :
\begin{equation}
\mathcal{L}_{\text{Huber}} =
\begin{cases}
\frac{1}{2}(\hat{S}-S)^2 & \text{if } |\hat{S}-S| \le \delta \\
\delta\left(|\hat{S}-S| - \frac{\delta}{2}\right) & \text{otherwise}
\end{cases}
\label{eq:huber}
\end{equation}
with $\delta = 1.0$. Huber loss is combines the advantages of Mean Squared Error (MSE) and Mean Absolute Error (MAE). For small prediction gaps, it is quadratic, with moderate correction. On the other hand, for larger error, it changes to a linear penalty. This limited gradient stops the loss from exploding and is much more robust to clinical outliers.

\subsubsection{Head Correction Loss}
An additional correction mechanism is required to attribute the total score error back to the individual components. An error redistribution strategy dynamically adjusts the target for each head. First, a normalized weight is computed to quantify the relative error of each head \textit{h}. Next, a pseudo target is calculated based on the proportional contribution to the total error:
\begin{equation}
\begin{gathered}
w_h = \frac{|\hat{s}_h - s_h|}{\sum_{h'}|\hat{s}_{h'}-s_{h'}| + \epsilon}, \\[4pt]
\tilde{s}_h = \text{clip}\left(s_h - 0.1\,w_h(\hat{S}-S),\ 0,\ 2\right)
\end{gathered}
\label{eq:head_correction_target}
\end{equation}
where $\epsilon$ is a small constant for numerical stability. Clipping preserves the valid $\{0,1,2\}$ scoring range. The correction loss is then the mean squared error against these dynamically adjusted pseudo-targets:
\begin{equation}
\mathcal{L}_{\text{correction}} = \frac{1}{5}\sum_h (\hat{s}_h - \tilde{s}_h)^2
\label{eq:correction_loss}
\end{equation}

\subsubsection{Combined Loss}
The three losses are combined as:
\begin{equation}
\mathcal{L}_{\text{total}} = \mathcal{L}_{\text{focal}} + \lambda\left(\mathcal{L}_{\text{Huber}} + \mathcal{L}_{\text{correction}}\right), \quad \lambda = 0.2
\label{eq:combined_loss}
\end{equation}
The Huber and corrective losses together provide a common global consistency mechanism. The scale factor $\lambda$ is assigned a small value so that the network softly imposes aggregate coherence on the total Swede score without overriding the fine-grained, localized feature learning in each individual classification head.
\section{Experimental Results}
\label{sec:results}

\subsection{CIN Classification}
\begin{figure}[t]
    \centering
   
    \includegraphics[width=0.5\columnwidth, keepaspectratio]{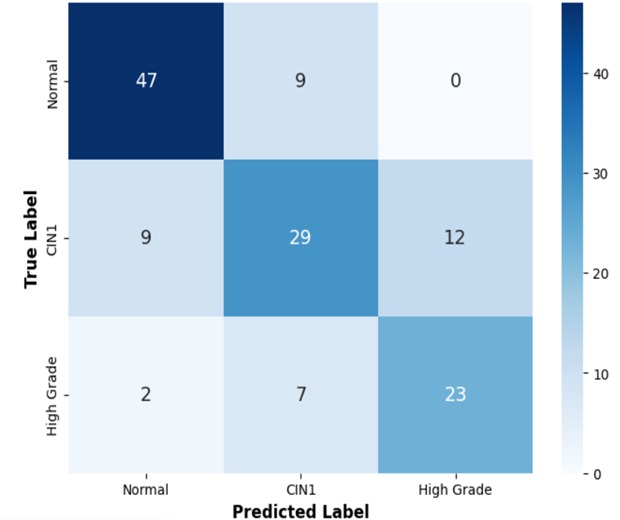}
    \caption{Confusion matrix illustrating the classification across the three diagnostic categories.}
    \label{fig:confusion_matrix}
\end{figure}

\begin{figure*}[tbh]
\centering
\subfloat[Aceto]{\includegraphics[width=0.19\textwidth]{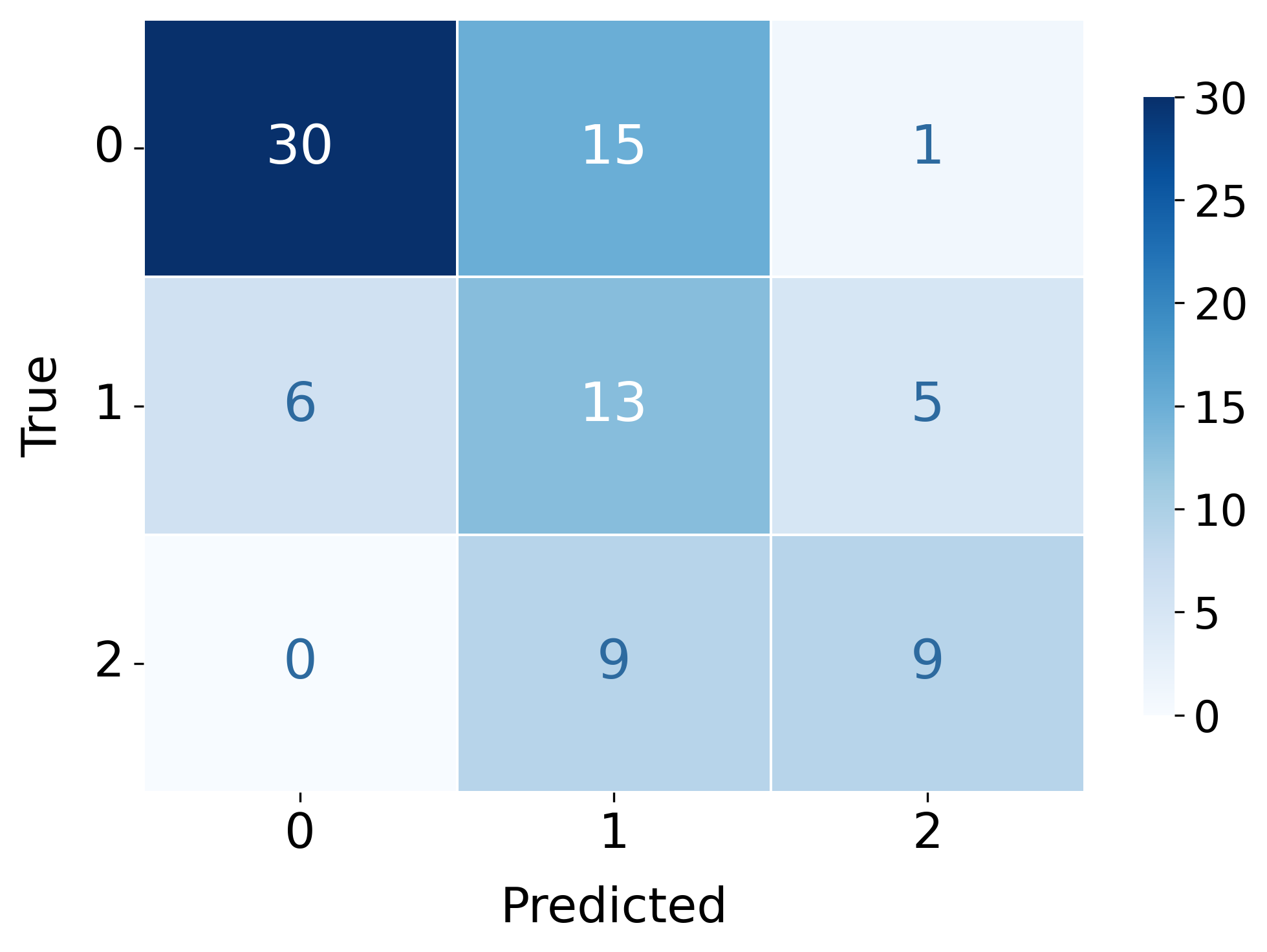}}
\hfill
\subfloat[Iodine]{\includegraphics[width=0.19\textwidth]{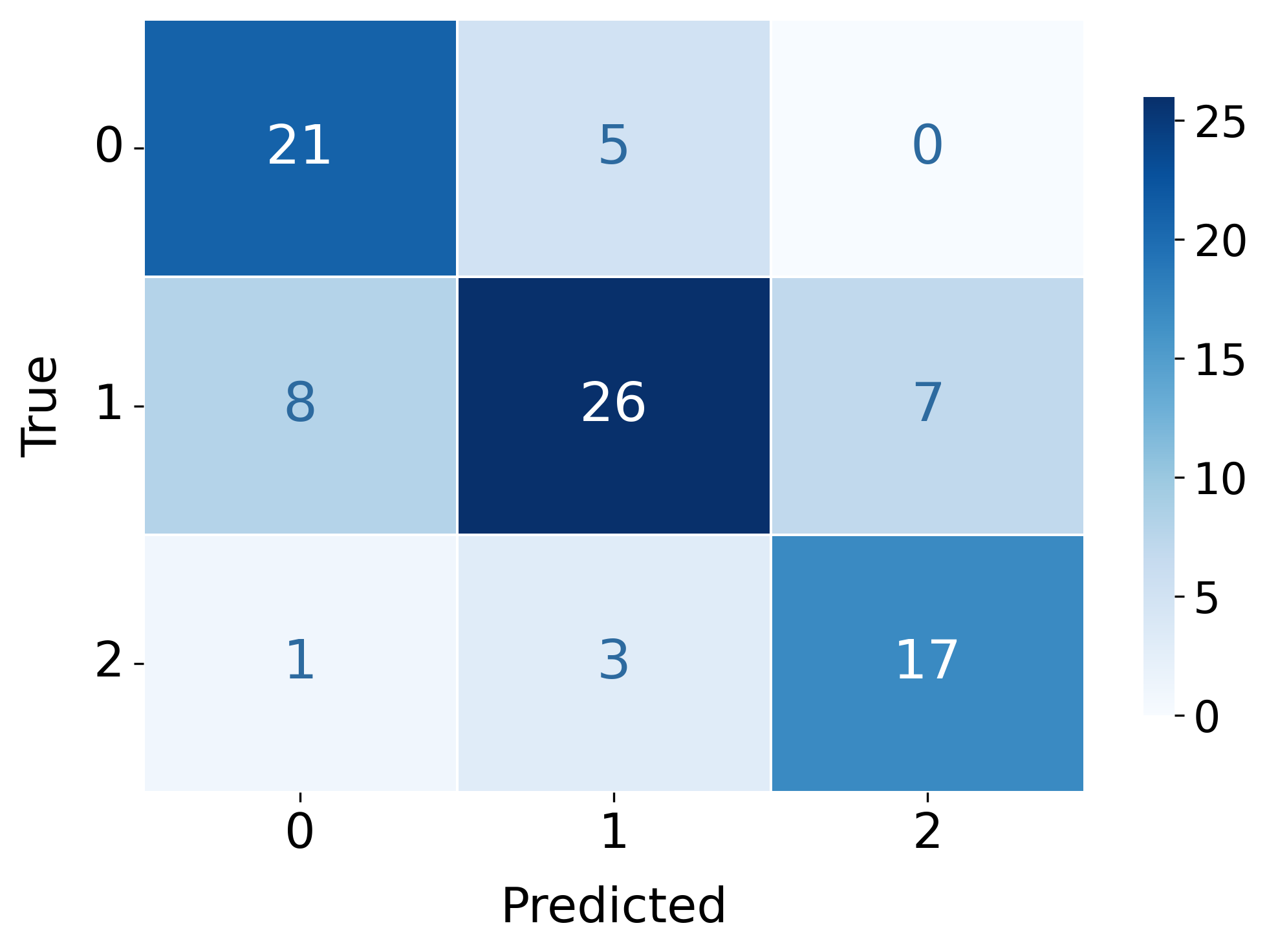}}
\hfill
\subfloat[Vessel]{\includegraphics[width=0.19\textwidth]{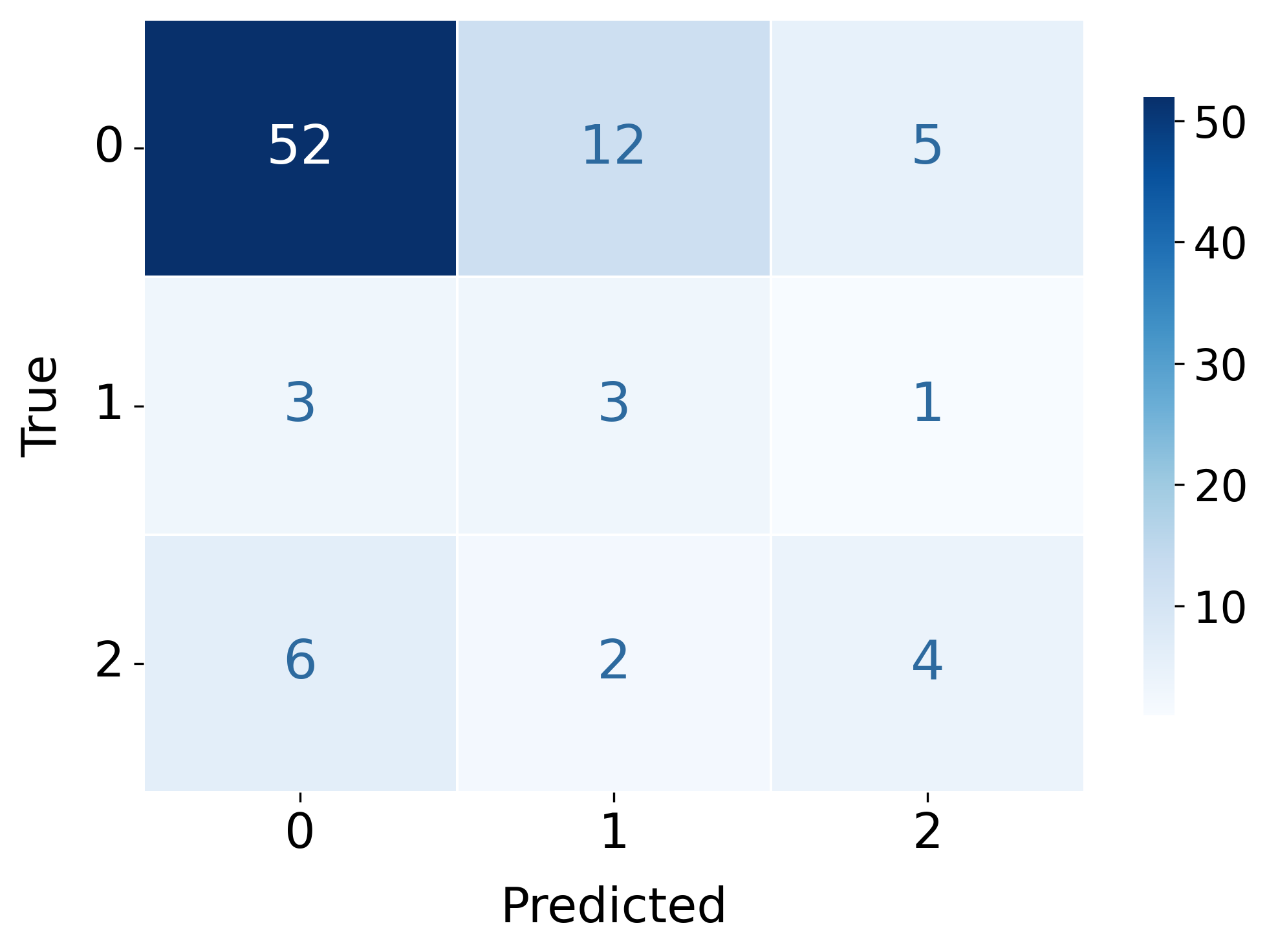}}
\hfill
\subfloat[Margin]{\includegraphics[width=0.19\textwidth]{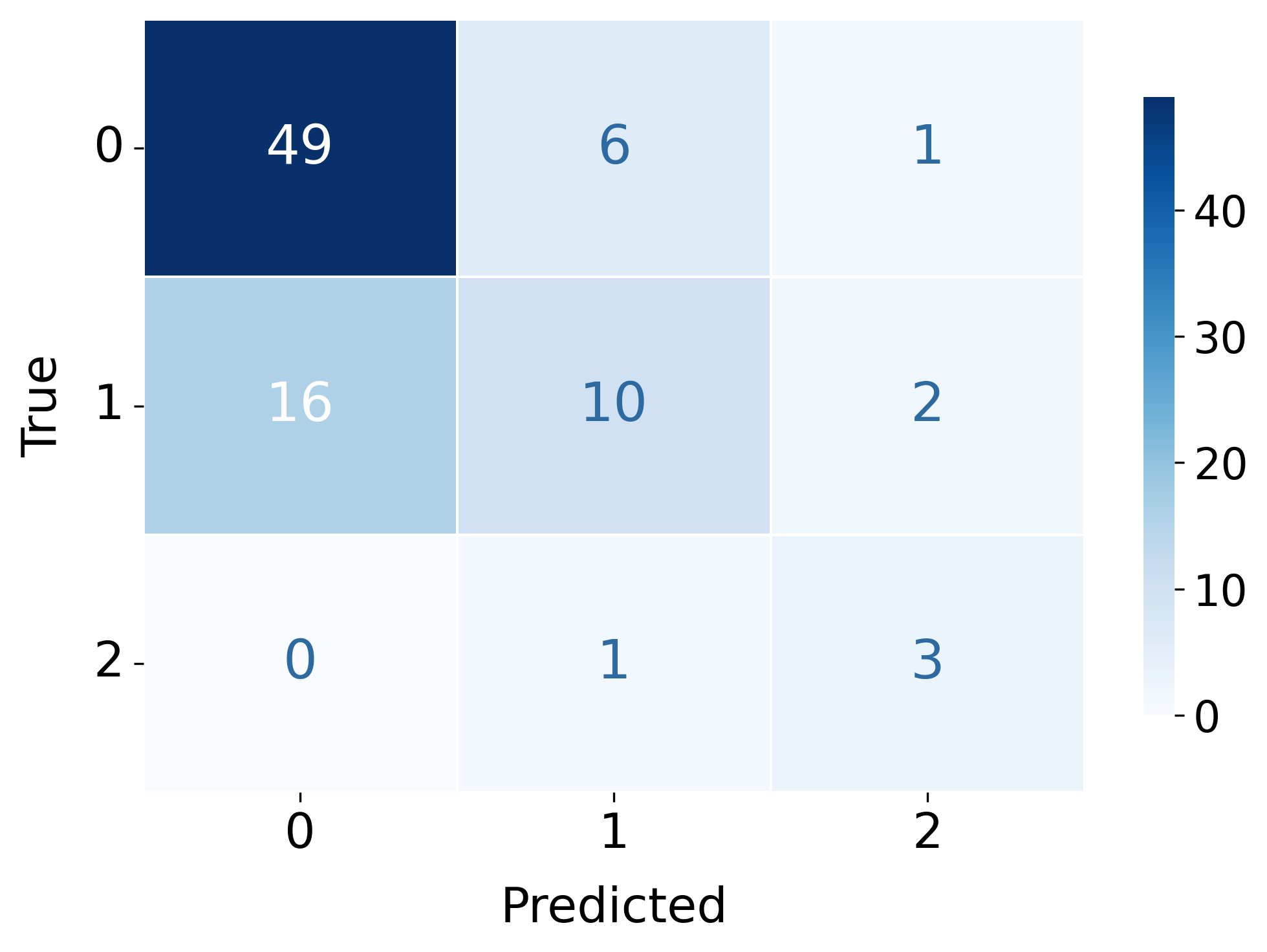}}
\hfill
\subfloat[Lesion Size]{\includegraphics[width=0.19\textwidth]{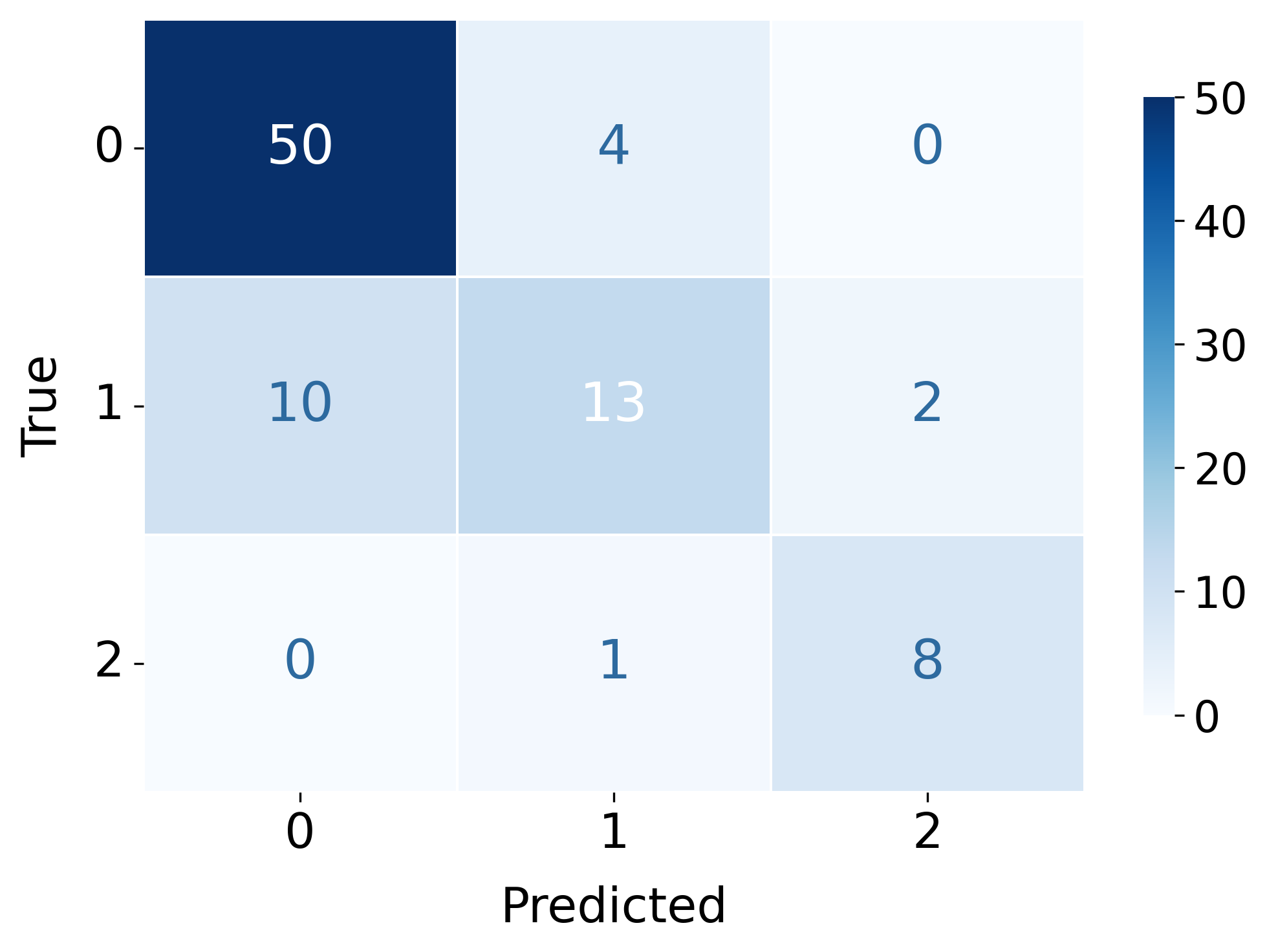}}

\caption{Confusion matrices for the proposed Swede Score Prediction Model across the five Swede score components: (a) Aceto Uptake, (b) Iodine Uptake, (c) Vessel Pattern, (d) Margin, and (e) Lesion Size.}
\label{fig:swede_confusion}
\end{figure*}

\begin{table*}[tbh]
\caption{Performance Metrics of Baseline and Proposed Models for CIN Classification}
\label{tab:cin_results}
\centering
\begin{tabular}{lcccccc}
\hline
\textbf{Model} & \textbf{Accuracy (\%)} & \textbf{Precision} & \textbf{Recall} & \textbf{F1 Score} & \textbf{AUC (\%)} & \textbf{Specificity (\%)} \\
\hline

EfficientNetB0 & 65.70 $\pm$ 2.39 & 0.65 $\pm$ 0.031 & 0.63 $\pm$ 0.028 & 0.63 $\pm$ 0.028 & 81.97 $\pm$ 1.22 & 84.11 $\pm$ 2.30 \\
DenseNet121 & 67.15 $\pm$ 1.23 & 0.67 $\pm$ 0.008 & 0.65 $\pm$ 0.026 & 0.65 $\pm$ 0.027 & 83.79 $\pm$ 0.49 & 81.98 $\pm$ 1.84 \\
Xception & 67.63 $\pm$ 3.26 & 0.67 $\pm$ 0.035 & 0.65 $\pm$ 0.047 & 0.65 $\pm$ 0.049 & 83.62 $\pm$ 1.90 & 80.89 $\pm$ 3.29 \\
InceptionV3 & 65.94 $\pm$ 2.05 & 0.65 $\pm$ 0.023 & 0.65 $\pm$ 0.013 & 0.65 $\pm$ 0.019 & 81.91 $\pm$ 0.45 & 83.16 $\pm$ 1.27 \\
ResNet50 & 66.43 $\pm$ 0.90 & 0.69 $\pm$ 0.006 & 0.63 $\pm$ 0.014 & 0.64 $\pm$ 0.014 & 81.97 $\pm$ 1.23 & 81.51 $\pm$ 1.54 \\

SWIN Transformer ~\cite{mohammed_swin_cnn} & 69.00 $\pm$ 3.84 & 0.67 $\pm$ 0.030 & 0.6402 $\pm$ 0.029 & 0.684 $\pm$ 0.033 & 80.71 $\pm$ 3.31 & 82.90 $\pm$ 2.88 \\
\textbf{Proposed Model} & \textbf{71.85 $\pm$ 3.80} & \textbf{0.705 $\pm$ 0.040} & \textbf{0.714 $\pm$ 0.039} & \textbf{0.706 $\pm$ 0.039} & \textbf{86.23 $\pm$ 2.46} & \textbf{85.70 $\pm$ 2.27} \\
\hline
\end{tabular}
\end{table*}

\begin{table*}[!t]
\caption{Comparison of Proposed Swede Score Component Prediction Model with Baseline Models}
\label{tab:swede_baseline}
\centering
\begin{tabular}{llccccc}
\hline
\textbf{Characteristic} & \textbf{Model} & \textbf{Accuracy} & \textbf{Macro-F1} & \textbf{Sensitivity} & \textbf{Specificity} & \textbf{AUC-ROC} \\
\hline
\multirow{5}{*}{Aceto Uptake}

& ResNet50 & 0.473 $\pm$ 0.039 & 0.300 $\pm$ 0.037 & 0.360 $\pm$ 0.021 & 0.651 $\pm$ 0.040 & 0.614 $\pm$ 0.034 \\
& ViT & 0.433 $\pm$ 0.015 & 0.301 $\pm$ 0.013 & 0.386 $\pm$ 0.010 & 0.709 $\pm$ 0.008 & 0.609 $\pm$ 0.018 \\
& EfficientNetB0 & 0.515 $\pm$ 0.023 & 0.349 $\pm$ 0.024 & 0.395 $\pm$ 0.028 & 0.667 $\pm$ 0.021 & 0.679 $\pm$ 0.014 \\
& \textbf{Proposed Model} & \textbf{0.600 $\pm$ 0.039} & \textbf{0.600 $\pm$ 0.122} & \textbf{0.594 $\pm$ 0.104} & \textbf{0.811 $\pm$ 0.088} & \textbf{0.766 $\pm$ 0.015} \\
\hline
\multirow{5}{*}{Iodine Uptake}

& ResNet50 & 0.500 $\pm$ 0.037 & 0.332 $\pm$ 0.070 & 0.394 $\pm$ 0.044 & 0.545 $\pm$ 0.050 & 0.665 $\pm$ 0.100 \\
& ViT & 0.463 $\pm$ 0.018 & 0.320 $\pm$ 0.011 & 0.386 $\pm$ 0.019 & 0.689 $\pm$ 0.014 & 0.647 $\pm$ 0.017 \\
& EfficientNetB0 & 0.648 $\pm$ 0.089 & 0.614 $\pm$ 0.100 & 0.617 $\pm$ 0.086 & 0.818 $\pm$ 0.019 & 0.792 $\pm$ 0.037 \\
& \textbf{Proposed Model} & \textbf{0.727 $\pm$ 0.040} & \textbf{0.733 $\pm$ 0.035} & \textbf{0.750 $\pm$ 0.093} & \textbf{0.859 $\pm$ 0.106} & \textbf{0.884 $\pm$ 0.007} \\
\hline
\multirow{5}{*}{Vessel Pattern}

& ResNet50 & 0.773 $\pm$ 0.009 & 0.291 $\pm$ 0.002 & 0.329 $\pm$ 0.004 & 0.868 $\pm$ 0.006 & 0.624 $\pm$ 0.026 \\
& ViT & 0.794 $\pm$ 0.012 & 0.333 $\pm$ 0.055 & 0.354 $\pm$ 0.034 & 0.684 $\pm$ 0.028 & 0.571 $\pm$ 0.034 \\
& EfficientNetB0 & 0.777 $\pm$ 0.011 & 0.292 $\pm$ 0.001 & 0.330 $\pm$ 0.005 & 0.872 $\pm$ 0.018 & 0.632 $\pm$ 0.024 \\
& \textbf{Proposed Model } & \textbf{0.670 $\pm$ 0.011} & \textbf{0.470 $\pm$ 0.278} & \textbf{0.503 $\pm$ 0.210} & \textbf{0.757 $\pm$ 0.280} & \textbf{0.757 $\pm$ 0.009} \\
\hline
\multirow{5}{*}{Margin}

& ResNet50 & 0.625 $\pm$ 0.009 & 0.336 $\pm$ 0.062 & 0.369 $\pm$ 0.040 & 0.739 $\pm$ 0.050 & 0.669 $\pm$ 0.036 \\
& ViT & 0.501 $\pm$ 0.007 & 0.311 $\pm$ 0.011 & 0.361 $\pm$ 0.004 & 0.692 $\pm$ 0.004 & 0.620 $\pm$ 0.009 \\
& EfficientNetB0 & 0.697 $\pm$ 0.028 & 0.412 $\pm$ 0.032 & 0.421 $\pm$ 0.027 & 0.806 $\pm$ 0.040 & 0.813 $\pm$ 0.007 \\
& \textbf{Proposed Model} & \textbf{0.705 $\pm$ 0.032} & \textbf{0.617 $\pm$ 0.185} & \textbf{0.663 $\pm$ 0.260} & \textbf{0.617 $\pm$ 0.289} & \textbf{0.846 $\pm$ 0.013} \\
\hline
\multirow{5}{*}{Lesion Size}

& ResNet50 & 0.625 $\pm$ 0.028 & 0.385 $\pm$ 0.068 & 0.409 $\pm$ 0.057 & 0.771 $\pm$ 0.060 & 0.700 $\pm$ 0.072 \\
& ViT & 0.496 $\pm$ 0.006 & 0.344 $\pm$ 0.020 & 0.412 $\pm$ 0.013 & 0.726 $\pm$ 0.004 & 0.678 $\pm$ 0.006 \\
& EfficientNetB0 & 0.640 $\pm$ 0.035 & 0.370 $\pm$ 0.036 & 0.395 $\pm$ 0.033 & 0.791 $\pm$ 0.034 & 0.772 $\pm$ 0.024 \\
& \textbf{Proposed Model} & \textbf{0.807 $\pm$ 0.062} & \textbf{0.773 $\pm$ 0.147} & \textbf{0.778 $\pm$ 0.215} & \textbf{0.864 $\pm$ 0.116} & \textbf{0.880 $\pm$ 0.012} \\
\hline
\end{tabular}
\end{table*}

\begin{table*}[!t]
\caption{Ablation Study Comparing Proposed Swede Score Component Prediction Model Against Single-Stream and No-Custom-Loss Variants}
\label{tab:swede_ablation}
\centering
\begin{tabular}{l  c c c  c c c c c}
\hline 
\textbf{Characteristic} & \textbf{Dual-Stream} & \textbf{Custom Loss} & \textbf{Accuracy} & \textbf{Sensitivity} & \textbf{Specificity} & \textbf{AUC-ROC} & \textbf{F1 Score} \\
\hline 
\multirow{3}{*}{Aceto Uptake}
 & & \checkmark & \textbf{0.730} & 0.569 & \textbf{0.856} & 0.762 & 0.538 \\
 & \checkmark & & 0.550 & 0.481 & 0.740 & 0.689 & 0.491 \\
 & \checkmark & \checkmark & {0.600} & \textbf{0.594} & {0.811} & \textbf{0.766} & \textbf{0.600} \\
\hline
\multirow{3}{*}{Iodine Uptake}
& & \checkmark & 0.662 & 0.643 & 0.806 & 0.815 & 0.595 \\
 & \checkmark & & 0.613 & 0.630 & 0.810 & 0.797 & 0.614 \\
& \checkmark & \checkmark & \textbf{0.727} & \textbf{0.750} & \textbf{0.859} & \textbf{0.884} & \textbf{0.733} \\
\hline
\multirow{3}{*}{Vessel Pattern}
& & \checkmark & 0.482 & 0.447 & 0.741 & 0.648 & 0.349 \\
 & \checkmark & & 0.662 & 0.578 & 0.768 & 0.792 & 0.456 \\
 & \checkmark & \checkmark & \textbf{0.670} & \textbf{0.503} & \textbf{0.757} & \textbf{0.757} & \textbf{0.470} \\
\hline
\multirow{3}{*}{Margin}
& & \checkmark & 0.518 & 0.331 & 0.692 & 0.589 & 0.295 \\
& \checkmark & & 0.675 & 0.663 & \textbf{0.794} & 0.785 & 0.579 \\
& \checkmark & \checkmark & \textbf{0.705} & \textbf{0.663} & {0.617} & \textbf{0.846} & \textbf{0.617} \\
\hline
\multirow{3}{*}{Lesion Size}
& & \checkmark & 0.686 & 0.650 & 0.847 & 0.825 & 0.620 \\
& \checkmark & & 0.725 & 0.664 & \textbf{0.887} & 0.822 & 0.635 \\
& \checkmark & \checkmark & \textbf{0.807} & \textbf{0.778} & {0.864} & \textbf{0.880} & \textbf{0.773} \\
\hline
\end{tabular}
\end{table*}

\begin{table}[!t]
\caption{Performance Metrics for Total Swede Score Prediction (0-10 Scale)}
\label{tab:total_swede_score}
\centering
\begin{tabular}{lccc}
\hline
\textbf{Configuration} & \textbf{MAE} & \textbf{MSE} & \textbf{RMSE} \\
\hline
Without custom loss & 1.809 $\pm$ 0.091 & 5.336 $\pm$ 0.627 & 2.306 $\pm$ 0.135 \\
\textbf{With custom loss} & \textbf{1.489 $\pm$ 0.021} & \textbf{4.312 $\pm$ 0.152} & \textbf{2.076 $\pm$ 0.037} \\
\hline
\end{tabular}
\end{table}

Table~\ref{tab:cin_results} compares the proposed CIN classification Model against seven conventional CNN baselines (MobileNetV3, EfficientNetB0, DenseNet121, Xception, InceptionV3, ResNet50, VGG16). Among conventional baselines, ResNet50, Xception, and DenseNet121 performed best. Despite being trained on a small dataset, the proposed model outperformed all baselines on all evaluated metrics. It achieved an accuracy of 71.85\% $\pm$ 3.80\% (95\% CI: 64.49\%--78.99\%), precision of 0.705 $\pm$ 0.040 (95\% CI: 0.628--0.780), recall of 0.714 $\pm$ 0.039 (95\% CI: 0.634--0.784), F1-score of 0.706 $\pm$ 0.039 (95\% CI: 0.629--0.777), and an AUC-ROC of 86.23\% $\pm$ 2.46\% (95\% CI: 81.13\%--90.71\%). These results and the CI values demonstrate the statistical reliability of the proposed architecture across different patient subsets. Additionally, we evaluated our dataset using the Swin Transformer architecture ~\cite{mohammed_swin_cnn}.

Confusion matrix (Fig \ref{fig:confusion_matrix}) analysis showed that the proposed model for CIN classification distinguished Normal from High Grade cases with near-zero confusion. Most misclassification occurred at the CIN1 boundary. This result is consistent with the established clinical reality that CIN1 represents a transitional, often ambiguous histological stage. Specificity was high for the proposed architecture, exceeding the best conventional baseline (EfficientNetB0, 84.11\%), indicating reliable rejection of true negative cases. This is an important property for avoiding unnecessary downstream referrals in resource-constrained clinical workflows.

\subsection{Swede Score Component Prediction}
The Swede score components exhibit substantial class imbalance across all five characteristics. This imbalance is a key factor in the relatively modest sensitivity and F1 scores despite strong AUC values. Table~\ref{tab:swede_baseline} compares the proposed model against the ViT, ResNet50, and EfficientNetB0 baselines across all five Swede score components under identical three-fold cross-validation. The proposed model achieves the best performance on the majority of metrics for every component. When compared to the highest-performing baseline model,EfficientNetB0, the proposed model achieved statistically significant improvements in AUC-ROC for Aceto Uptake (Proposed model:0.766; EfficientNetB0: 0.679, $p=0.0019$), Iodine Uptake (Proposed model: 0.884; EfficientNetB0: 0.792, $p=0.0456$) and Vessel Pattern (Proposed model:0.757; EfficientNetB0:0.632, $p=0.0063$). This significant performance gain was consistent across, Margin ($p=0.0293$), and Lesion Size ($p=0.0065$) as well. In both components, these improvements extend consistently across accuracy, recall, and specificity. 

The confusion matrices in Fig.~\ref{fig:swede_confusion} show that most misclassifications fall into the adjacent score category, indicating that the majority of errors are mild. Table~\ref{tab:total_swede_score} summarizes the total Swede Score prediction performance. As the total score can be within 0--10, we consider an MAE of 1.489 to be .

\subsection{Effect of the Custom Composite Loss}
Table~\ref{tab:swede_ablation} reports an ablation comparing Proposed Swede Score Prediction Model trained with and without the custom composite loss. The custom loss improves F1-score and AUC across all five components, with the largest surge observed for Lesion Size and Aceto Uptake. For Lesion Size, the F1-score improves from 0.635 to 0.773 and the AUC from 0.822 to 0.880. For Aceto Uptake, the F1-score improves from 0.491 to 0.600 and the AUC from 0.689 to 0.766. For Aceto Uptake and Iodine Uptake, every metric, including specificity, improves with the custom loss. As shown in Table~\ref{tab:total_swede_score}, the custom loss yields a decrease in MSE and RMSE, MAE being reduced by 17.69\%. 

\section{Discussion}
\label{sec:discussion}

Our dual-stream model fuses multimodal clinical features to reliably classify cervical tissues, prioritizing safety by minimizing critical false negatives. Extreme misclassifications between severe lesions and normal tissue are highly rare. While subtle morphological transitions cause minor overlap between adjacent pathological stages, the model maintains balanced sensitivity and high specificity. This strictly prevents dangerous under-referrals, confirming its efficacy as a safe, automated diagnostic tool.

For Swede score predictions, errors remain clinically conservative; confusion between Score 0 and Score 2 is rare overall and zero for Lesion Size. Although baselines show higher raw accuracy for features like Vessel Pattern due to majority-class bias, our model achieves superior discriminative capacity with higher specificity and AUC-ROC across all five components. These gains are directly driven by our custom composite loss, which mitigates class imbalance and substantially improves MAE, RMSE. Thus the total swede score predicted can be reliably mapped for CIN grading.

\section{Conclusion}
\label{sec:conclusion}
This study introduces a novel multi-center diverse colposcopy dataset and proposes a dual cross-attention framework for automated colposcopy image analysis. Featuring a three-class CIN grader and a multi-head Swede score predictor powered by a custom composite loss, our models consistently outperform established CNNs. Despite the challenges posed by a small, severely imbalanced, and heterogeneous multi-source dataset, the proposed framework achieves robust diagnostic accuracy, proving highly effective for resource-constrained clinical environments. Our future work aims to address two key limitations. Firstly, to mitigate the Swede sub-score class imbalance and improve minority-class reliability, we plan to combine targeted data collection with resampling and loss reweighting. Secondly, to rigorously validate generalization across different patient populations and imaging equipment, we aim to transition from pooled training to evaluating the model on unseen healthcare centers.
\balance
\bibliographystyle{IEEEtran}
\bibliography{references}

@misc{who_cervical_cancer,
  author       = {{World Health Organization}},
  title        = {Cervical Cancer Fact Sheet},
  year         = {2026},
  url          = {https://www.who.int/news-room/fact-sheets/detail/cervical-cancer},
  note         = {Accessed: 2026-08-21}
}

@article{uddin2023cervical,
  title={Cervical Cancer in Bangladesh},
  author={Uddin, AFM Kamal and Sumon, Mostafa Aziz and Pervin, Shahana and Sharmin, Farzana},
  journal={South Asian Journal of Cancer},
  volume={12},
  number={1},
  pages={36--38},
  year={2023},
  publisher={Thieme Medical Publishers},
  doi={10.1055/s-0043-1764202}
}

@article{ifcpc_nomenclature,
  author  = {J. Bornstein and J. Bentley and P. B{\"o}sze and F. Girardi and H. Haefner and M. Menton and M. Perrotta and W. Prendiville and P. Russell and M. Sideri and B. Strander and S. Tatti and A. Torn{\'e} and P. Walker},
  title   = {2011 Colposcopic Terminology of the {International Federation for Cervical Pathology and Colposcopy}},
  journal = {Obstetrics \& Gynecology},
  year    = {2012},
  volume  = {120},
  number  = {1},
  pages   = {166--172},
  doi     = {10.1097/AOG.0b013e318254f90c}
}

@article{swede_original,
  author  = {B. Strander and A. Ellstr{\"o}m-Andersson and S. Franz{\'e}n and I. Milsom and T. R{\aa}dberg},
  title   = {The Performance of a New Scoring System for Colposcopy in Detecting High-Grade Dysplasia in the Uterine Cervix},
  journal = {Acta Obstet. Gynecol. Scand.},
  year    = {2005},
  volume  = {84},
  number  = {10},
  pages   = {1013--1017},
  doi     = {10.1111/j.0001-6349.2005.00895.x}
}

@article{saini_colponet,
  author  = {S. K. Saini and V. Bansal and R. Kaur and M. Juneja},
  title   = {{ColpoNet} for Automated Cervical Cancer Screening Using Colposcopy Images},
  journal = {Mach. Vis. Appl.},
  year    = {2020},
  volume  = {31},
  pages   = {15},
  doi     = {10.1007/s00138-020-01063-8}
}

@article{yue_crcnn,
  author  = {Z. Yue and S. Ding and W. Zhao and H. Wang and J. Ma and Y. Zhang and Y. Zhang},
  title   = {Automatic {CIN} Grades Prediction of Sequential Cervigram Image Using {LSTM} with Multistate {CNN} Features},
  journal = {{IEEE} J. Biomed. Health Inform.},
  year    = {2020},
  volume  = {24},
  number  = {3},
  pages   = {844--854},
  doi     = {10.1109/JBHI.2019.2922682}
}

@article{chen_efficientnet_bigru,
  author  = {X. Chen and X. Pu and Z. Chen and L. Li and K.-N. Zhao and H. Liu and H. Zhu},
  title   = {Application of {EfficientNet-B0} and {GRU}-Based Deep Learning on Classifying the Colposcopy Diagnosis of Precancerous Cervical Lesions},
  journal = {Cancer Med.},
  year    = {2023},
  volume  = {12},
  pages   = {8690--8699},
  doi     = {10.1002/cam4.5581}
}

@article{mohammed_swin_cnn,
  author  = {F. A. Mohammed and K. K. Tune and J. A. Mohammed and T. A. Wassu and S. Muhie},
  title   = {Early Cervical Cancer Diagnosis with {SWIN}-Transformer and Convolutional Neural Networks},
  journal = {Diagnostics},
  year    = {2024},
  volume  = {14},
  number  = {20},
  pages   = {2286},
  doi     = {10.3390/diagnostics14202286}
}

@article{cin_last_comparison,
  author  = {B.-J. Cho and Y. J. Choi and M.-J. Lee and J. H. Kim and G.-H. Son and S.-H. Park and H.-B. Kim and Y.-J. Joo and H.-Y. Cho and M. S. Kyung and Y.-H. Park and B. S. Kang and S. Y. Hur and S. Lee and S. T. Park},
  title   = {Classification of Cervical Neoplasms on Colposcopic Photography Using Deep Learning},
  journal = {Sci. Rep.},
  year    = {2020},
  volume  = {10},
  pages   = {13652},
  doi     = {10.1038/s41598-020-70490-4}
}

@article{alhejri_vit_cytology,
  author  = {A. M. Al-Hejri and R. M. Al-Tam and A. H. Sable and B. Almuhaya and S. S. Alshamrani and K. M. Alshmrany},
  title   = {A Hybrid Vision Transformer with Ensemble {CNN} Framework for Cervical Cancer Diagnosis},
  journal = {BMC Med. Inform. Decis. Mak.},
  year    = {2025},
  volume  = {25},
  pages   = {411},
  doi     = {10.1186/s12911-025-03250-x}
}

@article{kalbhor_deepcervicancer,
  author  = {M. Kalbhor and S. Shinde and S. Lahade and T. Choudhury},
  title   = {{DeepCerviCancer} -- Deep Learning-Based Cervical Image Classification Using Colposcopy and Cytology Images},
  journal = {EAI Endorsed Trans. Pervasive Health Technol.},
  year    = {2023},
  volume  = {9},
  doi     = {10.4108/eetpht.9.3473}
}

@article{skerrett_pocket_colposcope,
  author  = {E. Skerrett and Z. Miao and M. N. Asiedu and M. Richards and B. Crouch and G. Sapiro and Q. Qiu and N. Ramanujam},
  title   = {Multicontrast Pocket Colposcopy Cervical Cancer Diagnostic Algorithm for Referral Populations},
  journal = {BME Front.},
  year    = {2022},
  volume  = {2022},
  pages   = {9823184},
  doi     = {10.34133/2022/9823184}
}

@article{fang_shufflenet,
  author  = {S. Fang and J. Yang and M. Wang and C. Liu and S. Liu},
  title   = {An Improved Image Classification Method for Cervical Precancerous Lesions Based on {ShuffleNet}},
  journal = {Comput. Intell. Neurosci.},
  year    = {2022},
  volume  = {2022},
  pages   = {9675628},
  doi     = {10.1155/2022/9675628}
}

@article{ren_annocerv,
  author  = {D. A. Minciun{\u{a}} and D. G. Socolov and A. Sz{\H{o}}cs and D. Ivanov and T. G{\^\i}sc{\u{a}} and V. Nechifor and S. Budai and A. G{\'a}l and {\'A}. B{\'a}lint and R. Socolov and D. Iclanzan},
  title   = {{AnnoCerv}: A New Dataset for Feature-Driven and Image-Based Automated Colposcopy Analysis},
  journal = {Acta. U. Sapien. Inform.},
  year    = {2023},
  volume  = {15},
  number  = {2},
  pages   = {306--329},
  doi     = {10.2478/ausi-2023-0019}
}

@inproceedings{cbam_woo,
  author    = {S. Woo and J. Park and J.-Y. Lee and I. S. Kweon},
  title     = {{CBAM}: Convolutional Block Attention Module},
  booktitle = {Proc. ECCV},
  pages     = {3--19},
  year      = {2018}
}

@inproceedings{cui_ens_2019,
  author    = {Y. Cui and M. Jia and T.-Y. Lin and Y. Song and S. Belongie},
  title     = {Class-Balanced Loss Based on Effective Number of Samples},
  booktitle = {Proc. IEEE CVPR},
  pages     = {9268--9277},
  year      = {2019},
  doi       = {10.1109/CVPR.2019.00949}
}

@article{huber_1964,
  author  = {P. J. Huber},
  title   = {Robust Estimation of a Location Parameter},
  journal = {Ann. Math. Statist.},
  volume  = {35},
  number  = {1},
  pages   = {73--101},
  year    = {1964}
}

@inproceedings{sechidis_multilabel2011,
  author    = {K. Sechidis and G. Tsoumakas and I. Vlahavas},
  title     = {On the Stratification of Multi-Label Data},
  booktitle = {ECML PKDD},
  series    = {Lect. Notes Comput. Sci.},
  volume    = {6913},
  pages     = {145--158},
  publisher = {Springer},
  year      = {2011}
}

@misc{iarc_image_bank,
  author       = {{International Agency for Research on Cancer (IARC)}},
  title        = {{IARC} Cervical Cancer Image Bank to Accelerate Innovation and Ensure the Quality of Artificial Intelligence Algorithms for Early Detection of Cervical Precancer and Cancer},
  howpublished = {\url{https://screening.iarc.fr/cervicalimagebank.php}},
  note         = {Accessed: 21 June 2024}
}
\end{document}